# Cross-Centroid Ripple Pattern for Facial Expression Recognition

Monu Verma, Prafulla Saxena, Santosh Kumar Vipparthi, Girdhari Singh

*Abstract—* **In this paper, we propose a new feature descriptor Cross-Centroid Ripple Pattern (CRIP) for facial expression recognition. CRIP encodes the transitional pattern of a facial expression by incorporating cross-centroid relationship between two ripples located at radius $r_1$ and $r_2$ respectively. These ripples are generated by dividing the local neighborhood region into subregions. Thus, CRIP has ability to preserve macro and micro structural variations in an extensive region, which enables it to deal with side views and spontaneous expressions. Furthermore, gradient information between cross centroid ripples provides strenght to captures prominent edge features in active patches: eyes, nose and mouth, that define the disparities between different facial expressions. Cross centroid information also provides robustness to irregular illumination. Moreover, CRIP utilizes the averaging behavior of pixels at subregions that yields robustness to deal with noisy conditions. The performance of proposed descriptor is evaluated on seven comprehensive expression datasets consisting of challenging conditions such as age, pose, ethnicity and illumination variations. The experimental results show that our descriptor consistently achieved better accuracy rate as compared to existing state-of-art approaches.**

## I. INTRODUCTION

Facial expressions exhibit human emotion, that is reflected directly on the face regions and gives cues about a person's feeling at that moment. Usually, expressions are more effective than verbal dialogue in order to perceive someone's feelings. Each expression represents a unique pattern on face based on the muscular variations at different facial regions. With rapid evolution of technology especially in computer vision and machine learning techniques, many state-of-the-art facial expression recognition (FER) systems have been proposed in the literature. Automatic facial expression recognition has various applications in different fields like depression analysis, e-learning, medical diagnosis, law enforcement and gaming, etc. In real world scenarios, facial images are captured from multiple angles with spontaneous interactions in various challenging environments such as illumination and noise variations. These factors make facial expression recognition a challenging task. Thus, there is a need for development of robust FER system, which can handle these challenging scenarios.

The concept of universal expression was given by Paul Ekman [1]. He classified expressions in six major categories namely: anger, disgust, fear, happy, sad and surprise. Furthermore, Facial Action Coding System [2] was built by Ekman and Friesen which standardized the recognition procedure. They achieved this by dividing the face into action units (AUs). These AUs are categorized according to movement of a facial muscle or set of muscles.

An automated FER system usually consists of three steps: preprocessing, feature extraction and expression classification. In preprocessing step, facial region is extracted or cropped from the image. This is important because presence of other elements that are not related to expressions (e.g. background in the image) can create confusion during recognition process. After preprocessing, features are extracted from processed images. This is the most important step in any facial expression recognition approach. A good feature extractor provides strength to FER system and improves accuracy. Feature extraction techniques can be divided into two categories namely predesigned and learning based. In predesigned techniques, handcrafted filters are designed to extract the features. They can be further divided into two categories namely geometric and appearance-based approaches. In geometry-based recognition, fiducial points are selected at active areas like eyebrows, eyes, nose and lips. With the help of these selected points, shape and distance between active areas are computed and then final features are extracted. There are many models available in the literature based on shape and size of active points. Overall performance of this type of models is dependent on selection of facial point with pin point accuracy, which is not easy to achieve because of dynamic nature of facial shape as it varies from expression to expression. In appearance-based approaches, features are extracted by applying hand-crafted filter over complete image or a part of image, called patch. Many filters such as Gabor Wavelets [3], Elastic Bunch Graph Matching [4], Local Binary Patterns (LBP) [5-7], Local Directional Pattern (LDP) [8], PHOG and LPQ [9] etc. are available for the same. These descriptors encode texture and gradient magnitude information to represent micro structure of a facial appearance. However, LBP and its variants are not

Monu Verma, Prafulla Saxena, Santosh Kumar Vipparthi and Girdhari Singh are with Department of Computer Science and Engineering, Malaviya National Institute of Technology, Jaipur, India (Email: monuverma.cv@gmail.com; prafulla1308@gmail.com; skvipparthi@mnit.ac.in; gsingh.cse@mnit.ac.in)

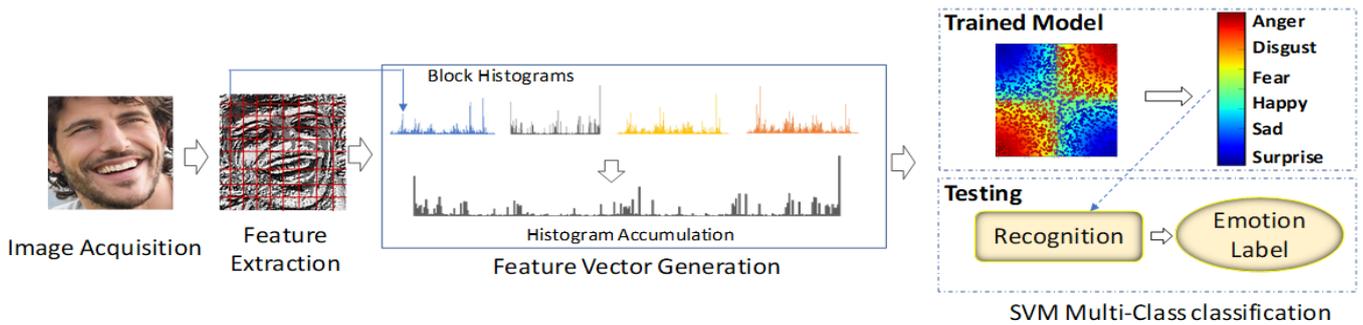

Fig. 1. Working block diagram of proposed FER system: The facial region is cropped from the original image and features are extracted by applying CRIP descriptor. Further, response map is divided into blocks to compute the feature vector. The feature vector is fed to the classifier for emotion recognition.

robust to noisy and multi-view pose conditions. After feature extraction, the next step, expression classification comes into play. It classifies the feature vectors into one of the six expression classes. Various classification models like Support Vector Machine (SVM) [10], K-Nearest Neighbor (KNN) [11], Hidden Marcov Models (HMM) [12], AdaBoost etc. are available for expressions classification. In the case of learning approaches, features are learned during training. Feature extraction and classification tasks are usually performed simultaneously by using some pipeline architecture.

In this paper, we propose Cross-Centroid Ripple Pattern (CRIP) for robust facial expression recognition that encodes the intensity variation using centroid values at adjacent ripples. Centroids of each ripple is calculated by establishing the average relationship between center pixels and its neighboring pixels located at radius $r_1$ and $r_2$ respectively. Thus, it embeds both micro and macro structure to represent the spatial variations and enhance discriminability between different expression classes. It improves performance of the descriptor in various pose conditions as it extracts deep edge responses from the face structure. The average relationship between the pixels provides a more comprehensive texture information and incorporates robustness to noise and image blur. Furthermore, gradient information between ripple centroids extract minute intensity variations and improves its ability to deal with illumination variations. The working flow of the proposed FER system is demonstrated in Fig. 1.

This paper is an extended version of the published conference paper: Quadrilateral Senary bit Pattern (QUEST) [13]. Similar to QUEST, we encode the edge patterns with the help of average information (centroids) of the local neighborhood by dividing the neighborhood into subregions. Mean of neighborhood provides a mutual behavioral knowledge of the included pixels. Therefore, it allows us to extract better transitional information between the pixels and enhance the discriminative capability of the descriptor.

QUEST partitioned the local neighborhood into two quadrilateral and then establish a trine relationship between two neighborhood and center pixel. However, QUEST descriptor is unable to capture inter radial transitional information, which also plays an essential role in defining a pattern of a structure in a local neighborhood. Therefore, CRIP descriptor incorporates information of the center pixel along with radius $r_1$, neighborhood pixels and inter-radial pixels of radius $r_1$ and $r_2$ to conserve macro and micro structural information of the facial appearance.

The remaining paper is structured as follows: In Section II, an overview of the existing state-of-the-art approaches is presented. In Section III, the proposed feature descriptor is described in detail with visual representation. Section IV represents the results of comparative study along with analysis outcome of the proposed descriptor. Finally, we conclude our work in Section V.

## II. RELATED WORK

In literature, FER techniques generalize in two categories: geometric and appearance based. Zang et al. [14] proposed a geometric based approach by using 34 facial point positions to extract the features. Another commonly used facial expression classification method is Facial Action Coding (FAC), where expressions are recognized by extracting AUs. Valstar et al. [15] introduced a new approach, where AUs are detected by extracting some particular facial points.

Appearance-based features are extracted by applying a filter or filter bank. These filters are applied on the whole face image or local regions of the face, which extracts the physical changes in facial appearance. There are many appearances-based techniques like Eigen faces [16], LDA [17], ICA [18], PCA [19], IDA [20], Gabor Wavelet [3] and LBP [5] proposed in literature, which extracts the local features to represent the physical appearance of the face. Local Binary Patterns (LBP) is the most acknowledged approach and has been effectively used in many domains. LBP used first-order derivative between the neighborhood and reference pixels to define the texture variation. The basic LBP descriptor molds each intensity value of an image to a new value by applying threshold condition between reference pixel to its neighboring pixel. LBP has gained significant popularity as invariant texture measure, as it has the tolerance capability to illumination changes and computational simplicity. Though, LBP performs well in extracting the micro information of edges, spots and other relevant features, by analyzing the intensity changes around the gray levels. But it is not robust to random noise and rotation variations. To overcome the issues of LBP, Jabid et al. [7] proposed a descriptor LDP, that utilize directional information instead of actual intensity value. To extract the directional numbers, LDP computed edge responses in eight directions by using krisch mask. Further, it extracts the three most leading directions indices and replace them by one binary and remaining with zeros. Use of edge responses showed successful performance in facial expression recognition as they are less sensitive in case of illumination changes and noise. However, LDP still face the problem of non- monotonic illumination variation. To improve the capability of feature descriptors, Ramirez et. al. [21] used the local directional numbers (LDN). LDN descriptor assigned six-bit code to each pixel value by extracting highest positive and negative edge responses. Further, Rivera et. al. [22] improved their directional descriptor's performance by adding texture information in it and proposed new descriptor LDTexP. LDTexP descriptor takes the advantages of both directional and intensity information of referencing neighborhood. Moreover, Ryu et. al. [23] proposed LDTerP, which encodes facial expressions by dividing the face into two parts as stable region and active regions. This multi-level representation provided a fine description of uniform and non-uniform areas. LDTerP encoded the texture information by using directional information with ternary pattern. Directional information reveals the edges pattern while ternary pattern described the disparities between excited and smooth facial regions. LDTerP used Robinson symmetric mask to extract the responses in four directions. Further, identified the two most dominating directions, jointly calculated the ternary patterns for both of the directions.

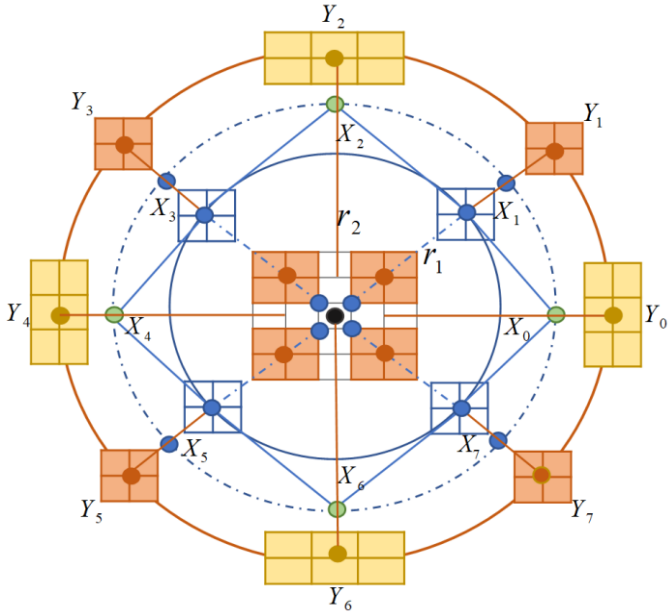

Fig. 2. Proposed CRIP descriptor: Centroids are computed by portioning the local region into sub-regions at radius $r_1$ and $r_2$. At each radius centroids $X(\eta)$ and $Y(\eta)$ are generated, where $\eta \in [0-7]$.

In recent years, deep learning techniques has performed very well in many applications such as: facial recognition, facial expression recognition, object detection, etc. In literature, various state-of-the-art CNN- based architectures such as VGG16 [24], VGG19 [24] and ResNet [25] have been proposed. These architectures have the ability to learn features itself with provided accurate labels. There are some networks available in literature which performing very well in FER, some of them are: Deep Belief Networks (DBN) [26], where each layer extract and learns the features from given image and predict the expression class. Khorrami et. al. [27] proposed a shallow CNN network to extract AU based features. Further, Li et al. [28] introduced an enhancing and cropping based CNN network to learn salient features from the AU region, by detecting landmarks. Lopes et al. [29] created big dataset by applied augmentation techniques and designed CNN based-network to detect the expressive features from the facial images. Ding et. al [30] proposed a two-phase network: first they trained a network for facial features and then forwarded updated neuron weights to the next network, to extract microlevel expression features. Other CNN based systems such as DTAGN [31], DTAGN-Joint [32], Inception [33], Spatiotemporal [34] and GCNet [35] have also achieved accelerated growth in the field of facial expression recognition.

## III. PROPOSED METHOD

In this paper, we propose a new feature descriptor named Cross-Centroid Rippled Pattern (CRIP) for facial expression recognition. CRIP is designed to encode the cross-centroid information, located at adjacent ripples to represent the spatial structure variation of the facial expressions. These ripples are generated by dividing the local neighborhood region into subregions and represented by $X_\eta$ and $Y_n \left(\eta \in \{0,1,...7\}\right)$ as shown in Fig. 2. The $X$ ripple is created by incorporating center pixel with its neighboring pixels, located at radius $r_1$. Similarly, $Y$ ripple is created by encapsulating the inter radial pixels, located at radius $r_1$ and $r_2$. Furthermore, centroid values are generated by computing local mean of all magnitude values in a particular subregion. Since, CRIP captures the averaging behavior of the subregion, that enhances noise robustness property of the proposed descriptor. Moreover, the CRIP efficiently encodes the macro and micro structure by embedding extensive information of radius $r_1$ and $r_2$ pixels. This enable proposed descriptor to become invariant view and posed facial expression. Final pattern is generated by establishing cross-centroid relationship between ripples, in the respective directions. Thereby, cross-centroid responses are able to capture the intensive and slight muscle distortions. Moreover, the CRIP pattern detects the deep and narrow edge responses, which reveal the salient facial features of the facial appearance that allows to handle irregular illumination. The properties of CRIP are as follows: (1) the proposed descriptor categorically reveals information of dipped and embossed muscles on face that is visible at the time of expression. Further, this enables identification of disparities between the intensity changes, (2) The CRIP extracts the robust spatial information embedding macro and micro structural information, that consequently yields better recognition accuracy, (3) The cross-centroid comparison between two adjacent ripples improve its capability to deal with irregular illumination changes, pose variations and noisy conditions.

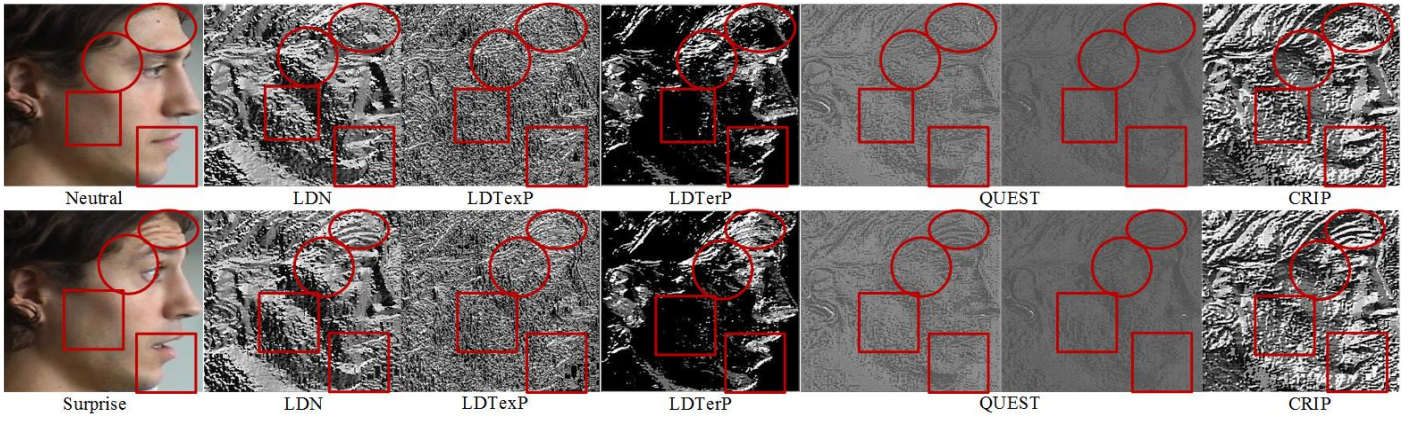

Fig. 3. The qualitative comparison between the feature maps generated by proposed CRIP and other state-of-the-art descriptors in side view image. Color boxes highlight the expressive regions of the facial appearance which plays major role in define the disparities between different class of the expressions.

To calculate the CRIP patterns, we divide the local neighborhood region into subregions and then compute the centroids to establish averaging relationship as shown in Fig. 2. Let $I(m,n)$ be a grayscale image of size $Q \times N$ where $m \in [1, Q]$, $n \in [1, N]$. If at each location $I_C$ in the image, p and q imply the total neighboring pixels located at corresponding radius $r_1$ and $r_2$ respectively, then the pattern is computed by using Eq. (1).

$$CRIP(c) = \sum_{\upsilon=0}^{p-1} f\left(Y(\upsilon) - X(\upsilon)\right) \times 2^{\upsilon} \tag{1}$$

The sign function $f(\cdot)$ is computed using Eq. (2).

$$f(t) = \begin{cases} 1 & t \geq 0 \\ 0 & else \end{cases} \tag{2}$$

The centroid values $X_\eta$ and $Y_n$ $\left(\eta \in \{0,1,...7\}\right)$ for radius $r_1$ and $r_2$ are calculated using [Eq. (3) – Eq. (6)].

$$X(\eta) = \frac{\left((k(\eta)-1) \times \sum_{i=\eta-l(\eta)-2}^{\eta-l(\eta)-1} I_{r_1, \text{mod}(i, p)}\right) + k(\eta) \times (I_C + I_{r_1, \eta-l(\eta)}) + \left(\sum_{i=\eta-l(\eta)+1}^{\eta-l(\eta)+2} I_{r_1, \text{mod}(i, p)}\right)}{4 \times k(\eta)} \tag{3}$$

$$Y(\eta) = \frac{\sum_{i=\eta+l(\eta)-1}^{\eta-l(\eta)+1} I_{r_1, \text{mod}(i, p)} + \sum_{i=2\eta-1}^{2\eta+1} I_{r_2, \text{mod}(i, q)}}{2\left(2 \times k(\eta) - \lfloor k(\eta)/2 \rfloor\right)} \tag{4}$$

$$k(\eta) = \text{mod}(\eta+1, 2) + 1 \tag{5}$$

$$l(\eta) = \text{mod}(\eta, 2) \tag{6}$$

### A. Comparison with Existing Descriptor

In the literature various state-of-the-art approaches have been proposed to extract salient features from the facial images. Recently, directional edge response-based approaches [7, 19-21] have achieved high classification accuracy. These methods have utilized the directional information, calculated by imposing the eight-directional compass mask (Sobel, kirsch and robinson). Moreover, they extract the directional index position to encode the feature in local neighborhood. Existing descriptors are able to describe the features of the still and posed frontal view expressions. However, in real life applications people express their emotions spontaneously, which can be recorded from multiple views. Thus, a feature descriptor which can also deal with the non-frontal views will yield better results for side-view and spontaneous expression recognition. Therefore, to improve the performance of FER systems, we proposed CRIP descriptor, which can extract expression features in such kind of situations. The comparison of different FER techniques can be summarized as follows.

1) The directional descriptors capture only extreme edge variations in the local neighborhood. This may lead to information loss in the facial region thereby degrading the performance of FER system. However, CRIP utilizes gradient information between centroids, located at different radials. Thus, CRIP Descriptor is able to extract efficient transitional pattern information to describe macro and micro structure variations which strengthens its capability to discriminate the expression classes in both side and frontal views as shown in Fig. 3 and Fig. 4 respectively.

2) In CRIP descriptor, gradient information of inter radial centroids, elicited from different sub regions, captures the dipped and embossed edges. Thus, it enhances the robustness to inter-class and intra-class variations as compared to LDTexP and LDTerP. The qualitative performance of the descriptors is depicted in Fig. 5, represents three expression images: two are from the anger emotion class and one is from the sad emotion class. We have extracted the response images along with histograms of active patches. The Euclidean distance is computed between the response feature vectors to validate the robustness of CRIP to discriminate between inter-class and intra-class variations. Thus, from the computed feature maps and distance measures, it is clear that CRIP is able to categorize anger expressions to the same class and sad expression to a different class.

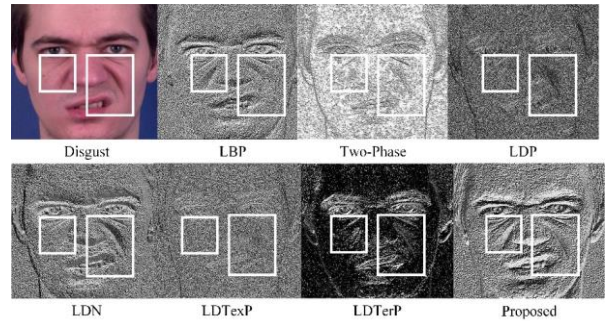

Fig. 4. An input anger expression and response feature maps encoded by a) LBP, b) Two-Phase, c) LDP, d) LDN, e) LDTexP, f) LDTerP and g) CRIP respectively.

3) The robustness of CRIP to deal with the illumination variations is compared to state-of-the-art techniques is depicted in Fig. 6. The cross-centroid relationship allows the CRIP to capture minute edge variations, that improves the feature maps quality in various lighting conditions over other state-of-the-art approaches.

4) CRIP utilizes averaging behavior of the pixels in local neighborhood to preserves edge variations in low resolution conditions (blur images). Fig. 7, demonstrates the response maps of happy expression in three different resolutions. It is visible that almost all existing methods loose prominent features in low resolution images as compare to CRIP, which degrades the performance of FER system.

5) It embeds extensive local neighborhood with radius $r_2$ instead of exclusive neighborhood with radius $r_1$ like QUEST, which elicits more pattern variations and extend the quality of response features.

Recently, convolution neural network (CNN) [24-25] based approaches have achieved high accuracy in FER. Conventional CNN models consist of a series of convolution, ReLu and pooling layers to create a hierarchical network. The network learns the relevant features of an expression and uses a softmax layer for classification. Although, deep learning models show remarkable performance in the field of image classification, but these networks also have some limitations such as 1) They require large datasets to get satisfactory results. 2) Predefined hyper-parameters play an important role in network learning. A minute change in hyper-parameter can affect the overall learning process which affects the performance indirectly. Parameter tuning in itself a difficult task and sometimes require numerous experiments to select optimized values. Therefore, the proposed designed feature CRIP is a relevant solution for automated facial expression recognition.

### B. Feature Vector & Classification

Feature response maps are divided into non-overlapping blocks of identical size for feature vector representation. The response feature map of size $u \times v$ is divided into $n$ equal-sized ($\eta \times \eta$) non-overlapping blocks $(K_1, K_2,...K_n)$. Thus, total number of blocks in a row and column are $R_\eta = \lceil u / \eta \rceil$, $C_\eta = \lceil v / \eta \rceil$ respectively. Final feature vector is generated by accumulating previously generated block level feature vectors. The feature vector $M_R$ is calculated by using Eq. (7).

$$M_R = \left\{ M(R, \lambda, \varphi) \right\}_{\varphi=1}^{n} \tag{7}$$

where $R$ is the generated feature map, $\lambda \in [0, 255]$ and $M(\cdot, \cdot, \cdot)$ is computed using [Eq. (8) - Eq. (10)].

$$M(R, \lambda, \varphi) = \sum_{i=\omega_1}^{\omega_1 + B} \sum_{i=\omega_2}^{\omega_2 + B} \sigma\left(M(i, j) - \lambda\right) \tag{8}$$

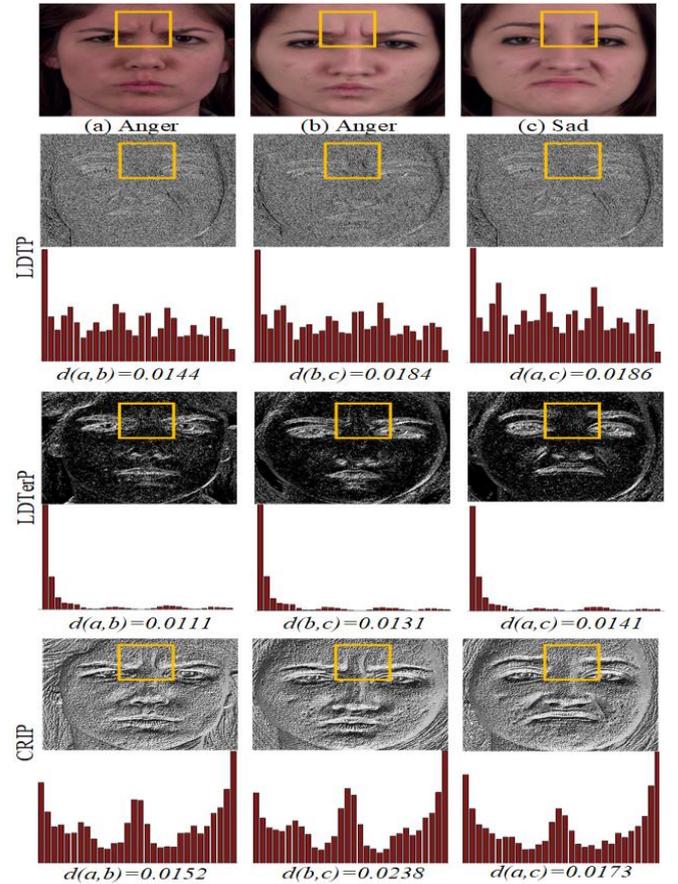

Fig. 5. The encoded response and feature vectors of three images with two emotion classes (anger and sad) are computed using LDTexP, LDTerP and CRIP respectively. Euclidean distance between (a) and (b) is lesser and (a) and (c) is larger for CRIP as compared to other descriptors.

$$\omega_1 = 1 + \left( \left\lfloor \varphi / R_j \right\rfloor \times B \right) \tag{9}$$

$$\omega_2 = \left( \mathrm{mod}(\varphi - 1, R_j) \times B \right) + 1 \tag{10}$$

where $\omega_1$ and $\omega_2$ is the row and column index for each block. We perform classification by using a support vector machine (SVM) [10]. SVM is a supervised machine learning technique, that, transforms a feature vector to a high dimensional plane by using linear or non-linear mapping. Furthermore, it executes the linear decision hyper-plane having maximal scope and performs binary classification to separate the feature vector into two classes in the dimensional space. Let $T = \{(A_i, B_i), i = 1, 2,....Z\}$ be a training set of labeled objects where $A_i \in \psi^n$ is the feature vector of each class and $B_i \in \{+1, -1\}$. A new test object is classified by using Eq. (11).

$$f(t) = sign\left( \sum_{i=1}^{Z} \lambda_i B_i \gamma(A_i, t) + b \right) \tag{11}$$

where $\lambda_i$ represents the Lagrange's multipliers of dual optimization problem, $\gamma$ is the kernel function and $b$ denote the bias. To perform multi-class classification, we adopt one-against-one coding scheme and use $^nC_2$ binary SVM classifiers, where $n$ is the total number of class label. The final decision is taken based on voting.

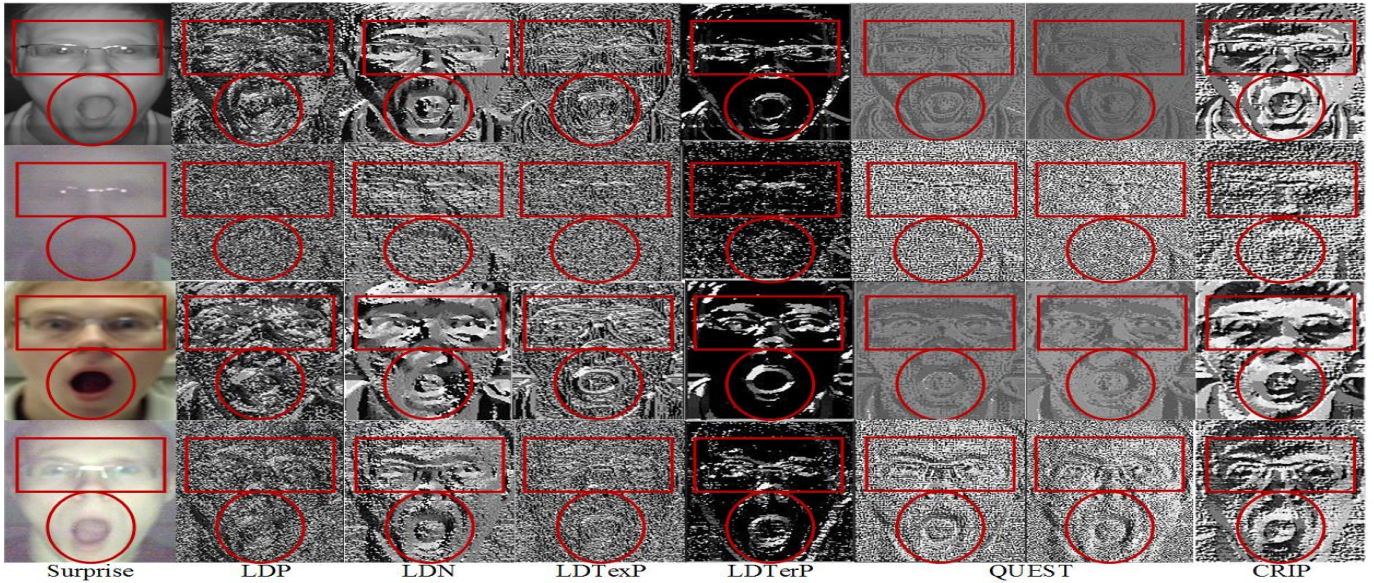

Fig. 6. The input images and coded response generated by applying state-of-art descriptors: LDP, LDN, LDTexP, LDTerP, QUEST and proposed CRIP. The images are captured using (a) NIR and (b- d) VIS cameras in various lighting conditions: dark, strong and week respectively.

## IV. EXPERIMENTAL RESULTS AND DISCUSSION

To evaluate the performance of CRIP encoding scheme, we execute 7 experiments on benchmark expression datasets MUG [36], MMI [37-38], Oulu-CASIA [39], ISED [40], AFEW [41,42], FER challenge [43] and Combined (MUG, MMI, ISED, AFEW). We tested CRIP descriptor under different conditions: time lapse, ethnicity and illumination variations. Moreover, we also validate our proposed approach with and without neutral images by creating datasets of six and seven expression classes respectively. Usually, existing approaches [21-23] perform manual detection of faces and then crop the facial region for further processing. However, we utilize the Viola Jones [44] detection algorithm to extract the faces and then normalize them into equal sized images. This procedure ensures simulation of a more realistic scenario for robust experimental evaluation. Furthermore, resultant feature maps are partitioned into equal sized blocks. The adaptability of equal sized blocks instead of selective block sizes for different databases, strengthens robustness of the FER system by reducing the overhead of parameter selection. This approach also ensures fair comparative analysis between proposed and existing descriptors. In our experimental setup, we have created our datasets by selecting three to five peak frames from each video sequence. Most of the existing approaches [8, 21-23] follow similar setup. In literature, to measure the performance of FER system, two validation schemes are used: person-dependent and person-independent cross-validation.

In person-dependent cross-validation, dataset is randomly partitioned into N categories. N-1 sets act as training set and remaining one set acts as a testing set. However, in person independent cross-validation, image sets are categorized according to the subjects. All expressions of a person/persons are not involved in training and utilized as a testing set. In this paper, we have validated performance of the proposed approach by utilizing both schemes.

**Person dependent scheme:** In person dependent (PD) scheme, we have arranged the training and testing set by dividing datasets into 80:20 respectively. Since, partition of train and test image set are done randomly, therefore, to make fair analysis of outcomes, we conduct the same experiments with 5 iterations and the average value is taken as final recognition rate. The recognition rate is computed by following equation.

$$Recognition\ Accuracy = \frac{Total\ no.\ of\ correctly\ predicted\ samples}{Total\ no.\ of\ samples} \times 100 \quad (12)$$

**Person independent scheme:** In person independent (PI) scheme, we have adopted two different PI methods: 10-fold subject independent and leave-on-subject-out. In 10-fold subject independent method, subjects are divided into 10 folds according to their ids. In our experiments, we have used 10-fold subject independent method for MMI and ISED datasets. Since, all the expressions of every subject is not available in these datasets, the 10-fold subject independent cross-validation is more suitable scheme. However, in leave-one-subject-out method, only one subject's expressions are involved in testing set and remaining all subject's expressions are used for training. For MUG and OULU-CASIA datasets, we have followed a leave-one-subject-out scheme.

TABLE I
RECOGNITION ACCURACY COMPARISON ON MUG DATASET

| Method | 6EXP(PD) | 7EXP(PD) | 6EXP(PI) | 7EXP(PI) |
|---|---|---|---|---|
| LBP [5] | 97.5 | 97.0 | 82.6 | 76.2 |
| Two-Phase [7] | 97.7 | 97.2 | 74.4 | 70.1 |
| LDP [8] | 98.1 | 97.4 | 82.8 | 78.7 |
| LDN [21] | 98.8 | 97.9 | 81.9 | 77.8 |
| LDTexP [22] | 98.1 | 98.0 | 82.0 | 78.7 |
| LDTerP [23] | 98.8 | 98.6 | 80.1 | 78.1 |
| QUEST [13] | 99.3 | 98.6 | 82.8 | 77.6 |
| VGG16 [24] | 99.6 | 98.8 | 85.1 | 84.6 |
| VGG19 [24] | 98.0 | 97.7 | 85.2 | 85.1 |
| ResNet50 [25] | 96.3 | 95.7 | 86.8 | 85.5 |
| **CRIP** | **99.6** | **99.0** | **83.5** | **78.9** |

*PD: Person Dependent, PI: Person Independent*

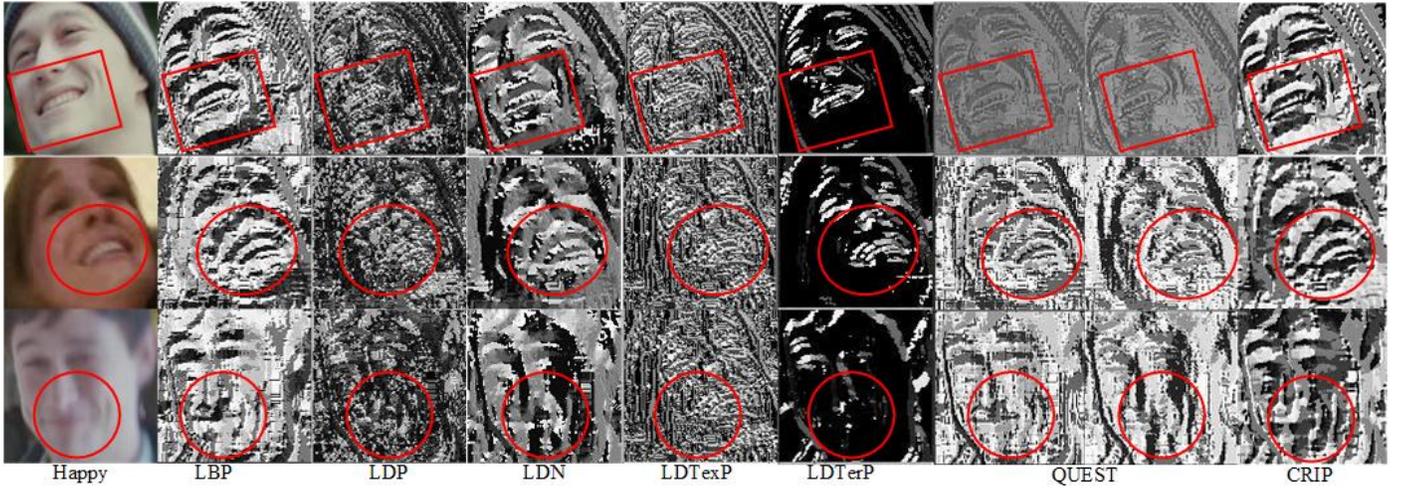

| Happy | LBP | LDP | LDN | LDTexP | LDTerP | QUEST | CRIP |

Fig. 7. Spontaneous happy emotion images, captured in three different resolution conditions and their coded response using LBP, LDP, LDN, LDTexP, LDTerP, QUEST and CRIP descriptors respectively.

TABLE II
RECOGNITION ACCURACY COMPARISON ON MMI DATASET

| Method | 6EXP(PD) | 7EXP(PD) | 6EXP(PI) | 7EXP(PI) |
|---|---|---|---|---|
| LBP [5] | 76.5 | 81.7 | 46.3 | 55.3 |
| Two-Phase [7] | 75.4 | 82.0 | 44.1 | 53.1 |
| LDP [8] | 80.5 | 84.0 | 45.7 | 55.8 |
| LDN [21] | 80.5 | 83.0 | 44.6 | 55.8 |
| LDTexP [22] | 83.4 | 86.0 | 46.5 | 56.1 |
| LDTerP [23] | 80.6 | 80.0 | 43.6 | 54.4 |
| QUEST [13] | 86.8 | 84.0 | 46.8 | 55.4 |
| VGG16 [24] | 83.9 | 89.2 | 66.5 | 70.9 |
| VGG19 [24] | 81.6 | 83.9 | 63.7 | 69.9 |
| ResNet50 [25] | 71.2 | 83.9 | 59.4 | 63.5 |
| **CRIP** | **87.6** | **86.6** | **47.9** | **56.6** |

*PD: Person Dependent, PI: Person Independent*

### A. Experiment on MUG Dataset

The multimedia understanding group built the MUG [36] dataset with high-resolution images. It consists of 86 subjects showing various expressions. The subjects contain 35 female and 512 males, all belongs to Caucasian origin aged between 20 to 35 years. We have taken 5 peak images from all the sequences of available subjects. images having 260 - anger, 250– disgust, 241 – fear, 255 – happy, 245– sad, 255 – surprise and 255 – neutral. For person dependent experiments, we arrange total 1761 For person independent experiments, we selected 1540 facial expression images (anger: 220, disgust: 220, fear: 220, happy: 220, sad: 220, surprise: 220, neutral: 220). We showed the recognition rates of our method and another methods over MUG dataset in Table I, from which we can state that our method outperforms the state-of-the-art methods for both PD and PI setup. The CRIP also outperforms VGG16, VGG19 and ResNet50 in PD experiments. Specifically, our proposed method yields performance improvement of 1.5%, 0.8%, 0.3 and 1.0%, 0.4%, 0.2% over LDTexP, LDTerP and QUEST for 6-class and 7-class problem respectively in PD experiments. Similarly, the proposed CRIP outperforms LDTexP, LDTerP and QUEST by 1.5%, 3.4%, 0.7% and 0.2%, 0.8%, 1.3% for 6-class and 7-class problem respectively in PI experiments.

### B. Experiment on MMI Dataset

The MMI [37-38] expression dataset has more than 2900 sample, which contains both videos and still images. Each sequence starts and ends with a neutral expression with a peak of particular expression in between the expression. The frontal view is captured in all images. There are multiple sessions for a person with and without glasses. MMI consists of 236 sessions recorded for 32 subjects. We selected 30 subjects frame sequence in our experiments. Each of the sequence categorized in one of expression class: anger, disgust, fear, happy, sad, surprise. We manually crop images of MMI dataset. For person dependent and person independent experiments, we have followed the similar setup and arranged total 561 images in which 69 – anger, 71 – disgust, 69 – fear, 84 – happy, 67 – sad, 75 – surprise and 126 – neutral. The recognition rates over MMI dataset for both 6-class and 7-class problems are shown in Table II. The results from Table II makes it clear that the proposed descriptor outperforms other existing handcrafted approaches for facial expression recognition. In particular, the proposed method improves the performance by 4.2%, 7.0%, 0.8% and 0.6%, 0.6%, 2.6% over LDTexP, LDTerP and QUEST for 6-class and 7-class problems respectively in PD experiments. Similarly, in PI experiments, the CRIP achieves 1.4%, 4.3%, 1.1% and 0.5%, 2.2%, 0.7% performance improvement over LDTexP, LDTerP and QUEST for 6-class and 7-class problems respectively. To delineate the class-wise recognition accuracy, we have further presented the confusion matrix for 6- and 7-class problems in Fig. 8 and Fig. 9 respectively.

|  | **True Label** | | | | | |
|---|---|---|---|---|---|---|
| **Predicated Label** | ANG | DIS | FEA | HAP | SAD | SUR |
| ANG | 13 | 1 | 0 | 0 | 0 | 0 |
| DIS | 1 | 12 | 0 | 0 | 1 | 0 |
| FEA | 0 | 1 | 12 | 0 | 0 | 1 |
| HAP | 1 | 0 | 0 | 16 | 0 | 0 |
| SAD | 1 | 0 | 1 | 0 | 11 | 0 |
| SUR | 1 | 0 | 1 | 1 | 0 | 12 |

Fig. 8. Confusion matrix of CRIP for 6-class expression classification in MUG dataset.

## C. Experiment on OULU-CASIA Dataset

OULU-CASIA [39] dataset consist of 80 subjects representing six basic expressions: anger, disgust, fear, happy, sad, surprise, aged between 23 to 58 years. All images in dataset are frontally faced. The expression images are captured by using NIR and VIS cameras. Furthermore, all images are taken in 3 different lighting conditions: strong, dark, and weak. In strong condition appropriate light is used, in weak condition only computer display light is used, while in dark condition images taken in near dark environment. Total 2880 video sequence are present in dataset, in which 480 image sequences belongs to each illumination condition. We only select 3 peak images from all Neutral expression is created by taking initial frames of the sequences when subjects are not started showing any emotion. For our experimental purpose, we arrange images in all the 3-illumination condition. The numbers of images we select in each expression as anger – 240, disgust – 240, fear – 240, happy – 240, surprise – 240 and neutral – 240. Same applied in both NIR and VIS camera images. To validate the robustness of CRIP, we have shown a qualitative comparison between the proposed descriptor and other state-of-the-art feature descriptors in Fig. 6. From Fig. 6, we can see that, almost all the methods perform well in case of normal lighting condition.

However, in the presence of strong and weak illumination condition, the proposed method is more conclusively able to extract the discriminative features as compared to other descriptors.

The quantitative results of the proposed method and other approaches over the NIR dataset for 6-class and 7-class problem is given through Table III - Table VI. The evaluated results clearly show that, the proposed descriptor outperforms other state-of-the-art approaches. More specifically, the proposed descriptor achieves performance gain of 0.8%, 0.9%, 0.6 and 1.0%, 0.5%, 0.3% for 6-class and 7-class problems as compared to LDTexP, LDTerP and QUEST respectively in PD setup. In PI setup, the CRIP secures 1.3%, 10.2%, 0.8% and 1.5%, 15.4%, 1.6% performance improvement over LDTexP, LDTerP and QUEST for 6-class and 7-class problems respectively. The results over VIS dataset are given in Table VII - Table X. In PD setup, the proposed method outperforms the LDTexP, LDTerP and QUEST by 2.6%, 1.3% 0.8% and 0.6%, 1.3%, 0.3 for 6-class and 7-class problems respectively. Similarly, in PI setup, the CRIP achieves 7.1%, 9.1%, 1.5% and 6.2%, 10.2%, 0.6% performance gain as compared to LDTexP, LDTerP and QUEST for 6-class and 7-class problems respectively.

## D. Experiment on ISED Dataset

The ISED [40] dataset includes 50 culturally diverse Indian subjects with 428 video sequences. Out of 50 subjects 29 are males and 21 are females with all aged between 18 to 22 years.

Fig. 9. Confusion matrix of CRIP for 7-class expression classification in MUG dataset

TABLE III

RECOGNITION ACCURACY COMPARISON ON OULU-NIR DATASET FOR 6-CLASS EXPRESSIONS IN PERSON DEPENDENT SETUP

| Method | Dark | Strong | Weak | Avg. |
|---|---|---|---|---|
| LBP [5] | 97.6 | 97.2 | 97.2 | 97.3 |
| Two-Phase [7] | 94.3 | 94.1 | 95.2 | 94.5 |
| LDP [8] | 96.6 | 97.5 | 97.9 | 97.3 |
| LDN [21] | 98.3 | 98.1 | 98.5 | 98.3 |
| LDTexP [22] | 98.1 | 98.0 | 98.2 | 98.1 |
| LDTerP [23] | 98.0 | 97.8 | 98.1 | 98.0 |
| QUEST [13] | 98.6 | 98.2 | 98.2 | 98.3 |
| **CRIP** | **98.6** | **98.4** | **98.7** | **98.9** |

TABLE IV

RECOGNITION ACCURACY COMPARISON ON OULU-NIR DATASET FOR 6-CLASS EXPRESSIONS IN PERSON INDEPENDENT SETUP

| Method | Dark | Strong | Weak | Avg. |
|---|---|---|---|---|
| LBP [5] | 64.09 | 67.09 | 63.26 | 64.8 |
| Two-Phase [7] | 47.36 | 48.51 | 44.93 | 46.9 |
| LDP [8] | 64.09 | 59.44 | 63.05 | 62.2 |
| LDN [21] | 63.05 | 65.48 | 64.65 | 64.4 |
| LDTexP [22] | 61.11 | 67.77 | 62.36 | 63.7 |
| LDTerP [23] | 56.04 | 57.08 | 51.31 | 54.8 |
| QUEST [13] | 62.77 | 65.69 | 64.37 | 64.3 |
| **CRIP** | **64.5** | **66.7** | **63.9** | **65.0** |

TABLE V

RECOGNITION ACCURACY COMPARISON ON OULU-NIR DATASET FOR 7-CLASS EXPRESSIONS IN PERSON DEPENDENT SETUP

| Method | Dark | Strong | Weak | Avg. |
|---|---|---|---|---|
| LBP [5] | 96.4 | 96.9 | 95.9 | 96.4 |
| Two-Phase [7] | 93.0 | 92.3 | 91.3 | 92.2 |
| LDP [8] | 96.0 | 97.7 | 97.7 | 97.1 |
| LDN [21] | 96.7 | 98.1 | 98.0 | 97.6 |
| LDTexP [22] | 97.8 | 97.7 | 97.1 | 97.5 |
| LDTerP [23] | 97.7 | 96.6 | 98.2 | 97.5 |
| QUEST [13] | 98.3 | 98.2 | 98.2 | 98.2 |
| **CRIP** | **98.2** | **98.5** | **98.5** | **98.5** |

TABLE VI

RECOGNITION ACCURACY COMPARISON ON OULU-NIR DATASET FOR 7-CLASS EXPRESSIONS IN PERSON INDEPENDENT SETUP

| Method | Dark | Strong | Weak | Avg. |
|---|---|---|---|---|
| LBP [5] | 63.6 | 65.7 | 61.8 | 63.7 |
| Two-Phase [7] | 45.4 | 46.3 | 44.5 | 45.4 |
| LDP [8] | 62.9 | 65.2 | 62.3 | 63.5 |
| LDN [21] | 63.2 | 64.8 | 60.6 | 62.8 |
| LDTexP [22] | 61.1 | 66.7 | 62.1 | 63.3 |
| LDTerP [23] | 40.6 | 55.8 | 51.9 | 49.4 |
| QUEST [13] | 61.6 | 65.8 | 62.2 | 63.2 |
| **CRIP** | **64.0** | **65.5** | **61.9** | **64.8** |

TABLE VII

RECOGNITION ACCURACY COMPARISON ON OULU-VIS DATASET FOR 6-CLASS EXPRESSIONS IN PERSON DEPENDENT SETUP

| Method | Dark | Strong | Weak | Avg. |
|---|---|---|---|---|
| LBP [5] | 94.1 | 96.3 | 96.1 | 95.5 |
| Two-Phase [7] | 80.3 | 87.8 | 90.0 | 86.0 |
| LDP [8] | 92.7 | 98.4 | 97.2 | 96.1 |
| LDN [21] | 94.3 | 98.5 | 96.0 | 96.2 |
| LDTexP [22] | 90.3 | 98.5 | 96.6 | 95.1 |
| LDTerP [23] | 93.9 | 98.3 | 97.2 | 96.4 |
| QUEST [13] | 94.5 | 98.5 | 97.9 | 96.9 |
| **CRIP** | **97.0** | **98.5** | **97.6** | **97.7** |

RECOGNITION ACCURACY COMPARISON ON OULU-VIS DATASET FOR 6-CLASS EXPRESSIONS IN PERSON INDEPENDENT SETUP

| Method | Dark | Strong | Weak | Avg. |
|---|---|---|---|---|
| LBP [5] | 55.9 | 75.3 | 57.7 | 62.9 |
| Two-Phase [7] | 31.0 | 55.8 | 38.4 | 41.8 |
| LDP [8] | 46.9 | 72.4 | 56.4 | 58.4 |
| LDN [21] | 53.8 | 72.2 | 58.6 | 61.6 |
| LDTexP [22] | 42.3 | 72.2 | 55.6 | 56.7 |
| LDTerP [23] | 43.2 | 68.5 | 52.7 | 54.8 |
| QUEST [13] | 53.2 | 75.9 | 57.5 | 62.3 |
| **CRIP** | **58.3** | **74.4** | **58.8** | **63.8** |

TABLE IX
RECOGNITION ACCURACY COMPARISON ON OULU-VIS DATASET FOR 7-CLASS EXPRESSIONS DEPENDENT SETUP

| Method | Dark | Strong | Weak | Avg. |
|---|---|---|---|---|
| LBP [5] | 90.1 | 93.3 | 94.1 | 92.5 |
| Two-Phase [7] | 86.2 | 87.0 | 89.4 | 87.5 |
| LDP [8] | 94.3 | 98.0 | 96.3 | 96.2 |
| LDN [21] | 95.3 | 97.8 | 96.7 | 96.6 |
| LDTexP [22] | 95.0 | 98.3 | 96.7 | 96.7 |
| LDTerP [23] | 92.4 | 98.8 | 96.8 | 96.0 |
| QUEST [13] | 94.9 | 99.1 | 97.2 | 97.0 |
| **CRIP** | **97.0** | **99.1** | **97.6** | **97.3** |

TABLE X
RECOGNITION ACCURACY COMPARISON ON OULU-VIS DATASET FOR 7-CLASS EXPRESSIONS IN PERSON INDEPENDENT SETUP

| Method | Dark | Strong | Weak | Avg. |
|---|---|---|---|---|
| LBP [5] | 57.4 | 65.7 | 57.3 | 60.1 |
| Two-Phase [7] | 28.1 | 53.9 | 39.2 | 40.4 |
| LDP [8] | 53.8 | 69.2 | 54.8 | 59.3 |
| LDN [21] | 53.8 | 70.9 | 59.3 | 61.3 |
| LDTexP [22] | 42.1 | 68.8 | 54.9 | 55.3 |
| LDTerP [23] | 40.6 | 64.5 | 48.7 | 51.3 |
| QUEST [13] | 54.7 | 70.7 | 57.3 | 60.9 |
| **CRIP** | **55.5** | **72.0** | **57.1** | **61.5** |

The dataset was created with spontaneous emotions, i.e., the subjects are not asked to pose particular expressions. These all spontaneous expressions were captured without subjects' awareness. Multiple sessions are recorded of same emotion with high resolution of 1920x1080. Expressions are categorized into 4 classes: disgust, happy, sad and fear. The neutral class created with the onset frames of these expressions sessions while subjects not started posing any expression. All captures are taken in weak illumination condition. Finally, we have selected 730 total image frames as: 234 – disgust, 294 – happy, 174 – sad, 189 – surprise and 295 – neutral. The proposed method gives more accurate results and extracts the discriminative features well as compare to other methods. The comparative results of our proposed method and other approaches given in Table XI. From Table XI, we can see that the proposed method achieves superior results as compared to other state-of-the-art approaches. Especially, it surpasses the accuracy of LDTexP, LDTerP and QUEST by 4.0%, 7.0%, 3.1% and 7.4%, 7.0%, 4.0% for 4-class and 5-class expression recognition problem in PD setup. Similarly, in PI setup, proposed descriptor attains 1.1%, 2.5%, 0.4% and 3.5%, 0.02%, 1.4% performance as compared to LDTexP, LDTerP and QUEST for 6-class and 7-class problems respectively Thus, it proves, that our method is effective even in recognizing the spontaneous expressions.

### E. AFEW

AFEW [41-42] contains 957 video clips of wild emotions around all age groups people. These videos are extracted from the different movies to encapsulate the real-world scenario for the dataset. The AFEW database comprises various real-world constraints: spontaneous facial expressions, head pose variations, occlusions, distinctive face resolution and illumination changes. In our setup, we have selected 10 to 15 most expressive images to validate the performance of the proposed method. In total, AFEW holds 10809 images, with seven basic expression classes: angry, disgust, fear, happy, sad, surprise and neutral. The comparative results of CRIP and other state-of-art approaches over AFEW are tabulated in Table XII. Specifically, our proposed method yields 8.0%, 1.9%, 1.2% and 6%, 1.3%, 0.9% more accuracy rates for 6- and 7- class expressions as compare to LDTexP, LDTerP and QUEST, respectively.

### F. Experiment on Combined Dataset

In this experiment we create a larger dataset by combining the MUG, MMI, ISED and AFEW datasets. By doing this we get a greater number of images to train the model and provides a diverse pool of subjects for designing and evaluating the FER system. The resultant dataset consists of 14242 images: 1856 -

TABLE XI
RECOGNITION ACCURACY COMPARISON ON ISED DATASET

| Method | 4EXP(PD) | 5EXP(PD) | 4EXP(PI) | 5EXP(PI) |
|---|---|---|---|---|
| LBP [5] | 88.9 | 83.0 | 63.3 | 59.0 |
| Two-Phase [7] | 78.4 | 72.1 | 58.6 | 50.9 |
| LDP [8] | 88.3 | 84.4 | 64.4 | 59.0 |
| LDN [21] | 90.6 | 85.6 | 64.2 | 55.8 |
| LDTexP [22] | 91.0 | 82.8 | 64.3 | 57.3 |
| LDTerP [23] | 88.0 | 83.2 | 62.8 | 55.8 |
| QUEST [13] | 91.9 | 86.2 | 64.9 | 57.9 |
| VGG16 [24] | 95.7 | 94.3 | 73.1 | 59.3 |
| VGG19 [24] | 92.9 | 94.3 | 70.2 | 69.2 |
| ResNet50 [25] | 90.2 | 94.3 | 68.1 | 63.5 |
| **CRIP** | **95.0** | **90.2** | **65.4** | **59.3** |

*PD: Person Dependent, PI: Person Independent*

TABLE XII
RECOGNITION ACCURACY COMPARISON ON AFEW DATASET

| Methods | 6EXP(PD) | 7EXP(PD) |
|---|---|---|
| LBP [5] | 80.5 | 81.0 |
| Two-Phase [7] | 57.0 | 52.6 |
| LDP [8] | 77.2 | 72.2 |
| LDN [21] | 81.5 | 79.9 |
| LDTexP [22] | 77.8 | 78.0 |
| LDTerP [23] | 83.9 | 82.7 |
| QUEST [13] | 84.6 | 83.1 |
| **CRIP** | **85.8** | **84.0** |

*PD: Person Dependent*

TABLE XIII
RECOGNITION ACCURACY COMPARISON ON COMBINED DATASET

| Methods | 6EXP(PD) | 7EXP(PD) |
|---|---|---|
| LBP [5] | 81.5 | 77.7 |
| Two-Phase [7] | 52.7 | 47.2 |
| LDP [8] | 73.4 | 70.4 |
| LDN [21] | 80.0 | 78.0 |
| LDTexP [22] | 79.7 | 75.6 |
| LDTerP [23] | 72.6 | 79.0 |
| QUEST [13] | 83.0 | 80.2 |
| **CRIP** | **84.5** | **82.5** |

*PD: Person Dependent*

| Method | 6EXP(Pb) | 7EXP(Pb) | 6EXP(Pr) | 7EXP(Pr) |
|--------|----------|----------|----------|----------|
| LBP [5] | 32.8 | 44.9 | 32.9 | 43.8 |
| Two-Phase [7] | 24.9 | 22.4 | 26.3 | 21.7 |
| LDP [8] | 41.2 | 35.3 | 40.9 | 42.4 |
| LDN [21] | 42.7 | 36.4 | 41.5 | 36.2 |
| LDTexP [22] | 43.2 | 37.9 | 42.4 | 39.3 |
| LDTerP [23] | 36.8 | 32.6 | 34.7 | 29.8 |
| QUEST [13] | 41.1 | 35.7 | 41.6 | 37.4 |
| **CRIP** | **51.8** | **46.1** | **50.5** | **45.8** |

*\* Pb: Public test set, Pr: Private test set*

anger, 1700 - disgust, 1497 - fear, 2761 - happy, 2265- sad, 1445 - surprise and 2718 - neutral. Table XIII represents the results for proposed method and other approaches. Table XIII represents the evaluated results for proposed descriptor and other approaches. The proposed method obtains 4.8%, 11.9%, 1.5% and 6.9%, 3.5%, 2.3% more recognition accuracy as compared to LDTexP, LDTerP and QUEST for 6-class and 7-class expression recognition respectively.

### G. FER 2013 Challenge

The FER challenge [43] was organized to promote the development of feature extraction and learning techniques for automated facial expression recognition. The dataset consists of grayscale images of 48x48 resolution. The images were partitioned into training set (28,709 images), public test set (3,589 images) and private test set (3,589 images). In our experiments, we have computed the results for 6-class and 7-class emotions over both the public and private test sets and the comparative results are shown in Table XIV. From Table XIV, we can see that the CRIP outperforms the existing state-of-the-art feature descriptors. More specifically, it attains 8.6%, 15.0%, 10.7% and 8.2%, 13.6%, 10.4% performance improvement as compared to LDTexP, LDTerP and QUEST for 6-class and 7-class problems in the public test set. Similarly, it outperforms the LDTexP, LDTerP and QUEST by 8.1%, 15.8%, 8.9 and 16%, 8.4% for 6-class and 7-class problems in the private test set.

## V. CONCLUSION

In this paper, we proposed a new feature descriptor Cross-Centroid Ripple Pattern, which encodes the image features by using inter radial ripples. These centroids are computed by establishing a relationship between neighboring subspaces in local neighborhood. The cross-centroid thresholding extracts the deep edge variations, which improves the capability to identify different expression belongs to different emotion classes. The cross-centroid relationship also enhances the robustness to pose and illumination variation, by highlighting the deep edge features. Moreover, average information of pixels in local neighborhood improves performance of CRIP in both frontal and side views. Experimental results show the effectiveness of the proposed method as it achieved better recognition rates as compared to other state-of-the-art approaches.

## REFERENCES

[1] P. Ekman, "Facial expression and emotion," *American psychologist*, vol. 48, no.4, pp. 384, 1993.

[2] P. Ekman, and W. V. Friesen, "Facial action coding system," 1977.

[3] D. Gabor, "Theory of communication," *J. Inst. Electr. Eng. III, Radio Commun. Eng.,* vol. 93, no. 26, pp. 429-425, 1946.

[4] L. Wiskott, N. Krüger and C. von der Malsburg, "Face recognition by elastic bunch graph matching," *IEEE Trans. Pattern Anal. Mach. Intell.,* vol. 19, no. 7, pp. 775-779, 1997.

[5] C. Shan, S. Gong and P. W. McOwan, "Facial expression recognition based on local binary patterns: A comprehensive study," *Image Vis. Comput.*, vol. 27, no. 6, pp. 803-816.

[6] G. Zhao and M. Pietikainen, "Dynamic texture recognition using local binary patterns with an application to facial expressions," *IEEE trans. on Pattern Anal. Mach. Intell*, vol. 29, no. 6, pp. 915-928, 2007.

[7] C. C. Lai and C. H. Ko, "Facial expression recognition based on two-stage features extraction.," *Optik-Int. J. Light Electron Optics*, vol. 125, no. 22, pp. 6678-6680, 2014.

[8] T. Jabid, M. Kabir and O. O. Chae, "Robust facial expression recognition based on local directional pattern," *J. ETRI*, vol. 32, no. 5, pp.784-794,2010.

[9] A. Dhall, R. Goecke, S. Lucey and T. Gedeon, "Static facial expression analysis in tough conditions: Data, evaluation protocol and benchmark," in *Proc. IEEE Int. Conf. of Comput. Vis.* pp. 2106-2112, 2011.

[10] C. Cortes and V. Vapnik, "Support-vector networks," *Machine learning,* vol. 20, no. 3, pp.273-297, 1995.

[11] W. Gu, C. Xiang, Y. V. Venkatesh, D. Huang and H. Lin, "Facial expression recognition using radial encoding of local Gabor features and classifier synthesis," *Pattern Recognit.*, vol. 45, no. 1, pp. 80-91, 2012.

[12] P. S. Aleksic and A. K. Katsaggelos, "Automatic facial expression recognition using facial animation parameters and multistream HMMs," *IEEE Trans. Inf. Forensics Security*, vol. 1, no. 1, pp.3-11, 2006.

[13] M. Verma, P. Saxena, S. K. Vippathi and G. Singh, "QUEST: Qudrilateral Senary bit Pattern for Facia Expression Recognition," *arXiv preprint arXiv: 1807.09154*, 2018.

[14] Z. Zhang, M. J. Lyons, M. Schuster, and S. Akamatsu, "Comparison between geometry-based and Gabor-wavelets-based facial expression recognition using multi-layer perceptron," in *Proc. 4th IEEE Int. Conf. Automatic Face Gesture. Recognit.*, pp. 454-459, 1998.

[15] M. Valstar, I. Patras, and M. Pantic, "Facial Action Unit Detection using Probabilistic Actively Learned Support Vector Machines on Tracked Facial Point Data," in *Proc. IEEE Conf. Comput. Vis. Pattern Recognit. Workshops,* vol. 3, pp. 76-84, 2005.

[16] M. Turk and A. Pentland, "Eigenfaces for recognition," *J. cognit. Neurosci.*, vol 3, no. 1, pp. 71-86, 1991.

[17] K. Etemad and R. Chellappa, "Discriminant analysis for recognition of human face images," *J. Opt. Soc. Amer. A, Opt. Image Sci.*, vol. 14, no. 8, pp. 1724–1733, 1997.

[18] M. S. Bartlett, J. R. Movellan and T. J. Sejnowski, "Face recognition by independent component analysis," *IEEE Trans. on neural networks*, vol. 13, no. 6, pp. 1450-1464, 2002.

[19] J. Yang, D. Zhang, A. F. Frangi and J. Y. Yang, "Two-dimensional PCA: a new approach to appearance-based face representation and recognition," *IEEE Trans. Pattern Anal. Mach. Intell.*, vol. 26, no. 1, pp. 131-137, 2004.

[20] Z. Nenadic, "Information discriminant analysis: Feature extraction with an information-theoretic objective," *IEEE Trans. Pattern Anal. Mach. Intell.*, vol. 29, no. 8, pp. 1394–1407, 2007.

[21] A. R. Rivera, J. R. Castillo and O. O. Chae, "Local directional number pattern for face analysis: Face and expression recognition," IEEE *Trans. Image Process.*, vol. 22, no. 5, pp.1740-1752, 2013.

[22] A. R. Rivera, J. R. Castillo and O. O. Chae, "Local directional texture pattern image descriptor," *Pattern Recognit. Letters*, vol. 51, pp.94-100, 2015.

[23] B. Ryu, A. R. Rivera, J. Kim and O. O. Chae, "Local directional ternary pattern for facial expression recognition," *IEEE Trans. Image Process.,* vol. 26, no. 12, pp. 6006-6018, 2017.

[24] K. Simonyan and A. Zisserman, "Very deep convolutional networks for large-scale image recognition," arXiv preprint arXiv:1409.1556, 2014.

[25] K. He, X. Zhang, S. Ren and J. Sun, "Deep residual learning for image recognition," in Proc. IEEE Conference on Computer Vision and Pattern Recognition, pp. 770-778, 2016.

[26] P. Liu, S. Han, Z. Meng and Y. Tong, "Facial expression recognition via a boosted deep belief network," In *Proc. IEEE Conf. Comput. Vis. Pattern Recognit.,* pp. 1805-1812, 2014.

[27] P. Khorrami, T. Paine and T. Huang, "Do deep neural networks learn facial action units when doing expression recognition," in *Proc. IEEE Int. Conf. Comput. Vis. Workshops*, pp. 19-27, 2015.

[28] W. Li, F. Abtahi, Z. Zhu and L. Yin, "EAC-Net: Deep Nets with Enhancing and Cropping for Facial Action Unit Detection," *IEEE Trans. Pattern Anal. Mach. Intell.,* 2018.

[29] A. T. Lopes, E. de Aguiar, A. F. De Souza and T. Oliveira-Santos, "Facial expression recognition with convolutional neural networks: coping with few data and the training sample order," *Pattern Recognit.*, vol. 61, pp. 610-628, 2017.

[30] H. Ding, S. K. Zhou and R. Chellappa, "Facenet2expnet: Regularizing a deep face recognition net for expression recognition," in *Proc. 12th IEEE Int. Conf. Automatic Face Gesture Recognit.,* pp. 118-126, 2017.

[31] H. Jung, S. Lee, S. Park, I. Lee, C. Ahn and J. Kim, "Deep temporal appearance-geometry network for facial expression recognition," in *arXiv preprint arXiv:1503.01532,* 2015.

[32] H. Jung, S. Lee, J. Yim, S. Park, and J. Kim, "Joint fine-tuning in deep neural networks for facial expression recognition," in *Proc. IEEE Int. Conf. Comput. Vis.,* pp. 2983-2991, 2015.

[33] A. Mollahosseini, D. Chan and M. H. Mahoor, "Going deeper in facial expression recognition using deep neural networks," in *Proc. IEEE Winter Conf. Applications Comput. Vis.,* pp. 1-10, 2016.

[34] K. Zhang, Y. Huang, Y. Du and L. Wang, "Facial expression recognition based on deep evolutional spatial-temporal networks," *IEEE Trans. Image Proc.,* vol. 26, no. 9, pp. 4193-4203, 2017.

[35] Y. Kim, B. Yoo, Y. Kwak, C. Choi and J. Kim, "Deep generative-contrastive networks for facial expression recognition," *arXiv preprint arXiv:1703.07140,* 2017.

[36] N. Aifanti, C. Papachristou and A. Delopoulos, "The MUG facial expression database*,"* in *Proc. 11th IEEE Int. Conf. Image Anal. Multimed. Interactive Workshop,* pp. 1-4, 2010.

[37] M. Pantic, M. Valstar, R. Rademaker and L. Maat, "Web-based database for facial expression analysis," in *Proc. IEEE Conf. Multimed. Expo.,* pp. 200-205, 2005.

[38] M. Valstar and M. Pantic, "Induced disgust, happiness and surprise: an addition to the mmi facial expression database," in *Proc. 3rd Int. Workshop. Emot. (Satellite of LREC),* pp. 65-70, 2010.

[39] G. Zhao, X. Huang, M. Taini, S.Z. Li and M. Pietikälnen, "Facial expression recognition from near-infrared videos," *Image Vis. Comput.*, vol. 29, no. 9, pp.607-619, 2011.

[40] S. L. Happy, P. Patnaik, A. Routray and R. Guha, "The Indian spontaneous expression database for emotion recognition," *IEEE Trans. Affect. Comput.*, vol. 8, no. 1, pp.131-142, 2017.

[41] A. Dhall, R. Goecke, S. Lucey and T. Gedeon, "Collecting Large, Richly Annotated Facial-Expression Databases from Movies," *IEEE Multimedia*, 2012

[42] A. Dhall, R. Goecke, S. Ghosh, J. Joshi, J. Hoey and T. Gedeon, "From individual to group-level emotion recognition: EmotiW 5.0.," in *Proc. 19th ACM Int. Conf. Multimodal Interaction,* pp. 524-528, 2017.

[43] Kaggle.com, "Challenges in Representation Learning: Facial Expression Recognition Challenge," [Online]: Available: https://www.kaggle.com/c/challenges-in-representation-learning-facial-expression-recognition-challenge/data.

[44] P. Viola and M. J. Jones, "Robust real-time face detection," *Int. J. Comput. Vis.,* vol. 57, no. 2, pp. 137-154, 2004.